\documentclass[letterpaper, 10 pt, conference]{ieeeconf}  

\IEEEoverridecommandlockouts                              
\usepackage{graphicx}
\usepackage{amsmath}
\usepackage{amsfonts}
\usepackage{amssymb}
\usepackage{tikz}
\usetikzlibrary{fit,positioning,quotes,arrows.meta}
\usepackage{algorithm}
\usepackage{booktabs}
\usepackage{mathtools}
\usepackage[noend]{algpseudocode}

\DeclareMathOperator*{\argmin}{arg\,min}
\DeclareMathOperator*{\argmax}{arg\,max}
\DeclareMathOperator{\E}{\mathbb{E}}

\DeclareMathOperator{\DKL}{\text{D}_\text{KL}}
\DeclareMathOperator{\F}{\Delta \text{F}_{\text{par}}}

\title{\LARGE \bf
An Information-theoretic On-line Learning Principle for Specialization in Hierarchical Decision-Making Systems
}

\author{Heinke Hihn, Sebastian Gottwald, and Daniel A. Braun
\thanks{This study was supported by the European Research Council (ERC-StG-2015 - ERC Starting Grant, Project ID: 678082, BRISC: Bounded Rationality in Sensorimotor Coordination) and Deutsche Forschungsgemeinschaft Emmy Noether-Programm BR 4164/1-1.}
\thanks{H. Hihn, S. Gottwald and D.A. Braun are with the Institute of Neural Information Processing at the Ulm University,
       Ulm University, 89073 Ulm, Germany.
        {Contact Mail: \tt\small heinke.hihn@uni-ulm.de}}%
}

\begin{document}
\maketitle
\thispagestyle{empty}
\pagestyle{empty}

\begin{abstract}
Information-theoretic bounded rationality describes utility-optimizing decision-makers whose limited information-processing capabilities are formalized by information constraints. One of the consequences of bounded rationality is that resource-limited decision-makers can join together to solve decision-making problems that are beyond the capabilities of each individual. Here, we study an information-theoretic principle that drives division of labor and specialization when decision-makers with information constraints are joined together. We devise an on-line learning rule of this principle that learns a partitioning of the problem space such that it can be solved by specialized linear policies. We demonstrate the approach for decision-making problems whose complexity exceeds the capabilities of individual decision-makers, but can be solved by combining the decision-makers optimally. The strength of the model is that it is abstract and principled, yet has direct applications in classification, regression, reinforcement learning and adaptive control.
\end{abstract}

\section{Introduction}
Intelligent learning systems are often formalized as decision-makers that learn probabilistic models of their environment and optimize utilities. Such utility functions can represent different classes of problems, such as classification, regression or reinforcement learning. To enable these agents to learn optimal policies, it is usually too costly to enumerate all possibilities and determine the expected utilities. Intelligent agents must instead invest their limited resources such that they optimally trade off utility versus processing costs \cite{Gershman2015}, which can be formalized in the framework of bounded rationality \cite{Simon1955}. The information-theoretic approach to bounded rationality \cite{Ortega2013} provides an abstract model to formalize how such agents behave in order to maximize utility within a given resource limit, where resources are quantified by information processing constraints \cite{Edward2014, McKelvey1995,Tishby2011,Wolpert2006}.

Intriguingly, the information-theoretic model of bounded rationality can also explain the emergence of hierarchies and abstractions, in particular when multiple bounded rational agents are involved in a decision-making process \cite{Genewein2015}. In this case an optimal arrangement of decision-makers leads to specialization of agents and an optimal division of labor, which can be exploited to reduce computational effort \cite{Gottwald2019}. 
Here, we introduce a novel gradient-based on-line learning paradigm for  hierarchical decision-making systems. Our method finds an optimal soft partitioning of the problem space by imposing information-theoretic constraints on both the coupling between expert selection and on the expert specialization. We argue that these constraints enforce an efficient division of labor in systems that are bounded. As an example, we apply our algorithm to systems that are limited in their representational power---in particular by assuming linear decision-makers that can be combined to solve problems that are too complex for each decision-maker alone.

The outline of this paper is as follows: first we give an introduction to bounded rationality, next we introduce our novel approach and demonstrate how it can be applied to classification, regression, non-linear control, and reinforcement learning. At last, we conclude.

\section{Background}
\subsection{Bounded Rational Decision Making}
\label{seq:br}
An important concept in decision-making is the Maximum  Utility principle \cite{VonNeumann2007}, where an agent always chooses an optimal action $a_s^* \in \mathcal{A}$ such that it maximizes their expected utility depending on the context $s \in \mathcal{S}$, i.e.
\begin{equation}
a^*_s = \argmax_a \mathbf{U}(s, a),
\end{equation}
where the utility is given by a function $\mathbf{U}(s, a)$ and the states are distributed according to a known and fixed distribution $p(s)$. Solving this optimization problem naively leads to exhaustive search over all possible $(a,s)$ pairs, which is in general a prohibitive strategy. One possible approach to this is studying decision-makers that have limited processing power, e.g. that have to act within a given time limit. Instead of finding an optimal strategy, a \textit{bounded rational decision-maker} optimally trades off expected utility and the processing costs required to adapt the system. In this study we consider the information-theoretic free energy principle \cite{Ortega2013, Ortega2015} of bounded rationality, where the decision-maker's behavior is described by a probability distribution $p(a|s)$ over actions $a$ given a particular state $s$ and the decision-maker‘s processing costs are given by the Kullback-Leibler divergence $\DKL(p(a|s)||p(a)) = \sum_{a}{p(a|s) \log{\frac{p(a|s)}{p(a)}}}$
between the agent‘s prior distribution over the actions $p(a)$ and the posterior policies $p(a|s)$.

We can model the decision-maker's processing power by defining an upper bound B on the average $\DKL$ the agent can maximally spend to adapt its prior behavior, which results in the following constrained optimization problem:
\begin{align}
\max_{p(a|s)} \sum_{s,a} p(s){p(a|s)\mathbf{U}(s, a)} \\ \text{ s.t. } \mathbb{E}_{p(s)}\left[\DKL(p(a|s)||p(a))\right] \leq \text{B}.
\end{align}
This constraint can be regarded as a regularization imposed on the forms the distributions can take. The resulting constrained optimization problem can be transformed into an unconstrained variational problem by introducing a Lagrange multiplier  $\beta \in \mathbb{R}^+$ that governs the trade-off between expected utility gain and information cost \cite{Ortega2013}:
\begin{equation}
\max_{p(a|s)} \mathbb{E}_{p(s|a)}\left[\mathbf{U}(s, a)\right] - \frac{1}{\beta}\mathbb{E}_{p(s)}\left[\DKL(p(a|s)||p(a))\right].
\label{eq:br_obj}
\end{equation}
For $\beta \rightarrow \infty$ the agent acts perfectly rational and for $\beta \rightarrow 0$ the agent can only act according to the prior. The optimal prior in this case is given by the marginal $p(a) = \sum_{s \in S}{p(s) p(a|s)}$. The solution to this optimization problem can be found by applying Blahut-Arimoto type algorithms, similar to Rate-Distortion Theory \cite{Blahut1972, Arimoto1972, Genewein2015}.

\subsubsection{Hierarchical Decision Making} 
\begin{figure}[t!]
\includegraphics[width=.5\textwidth, trim={0.03cm 25.5cm 6.15cm 0cm}, clip]{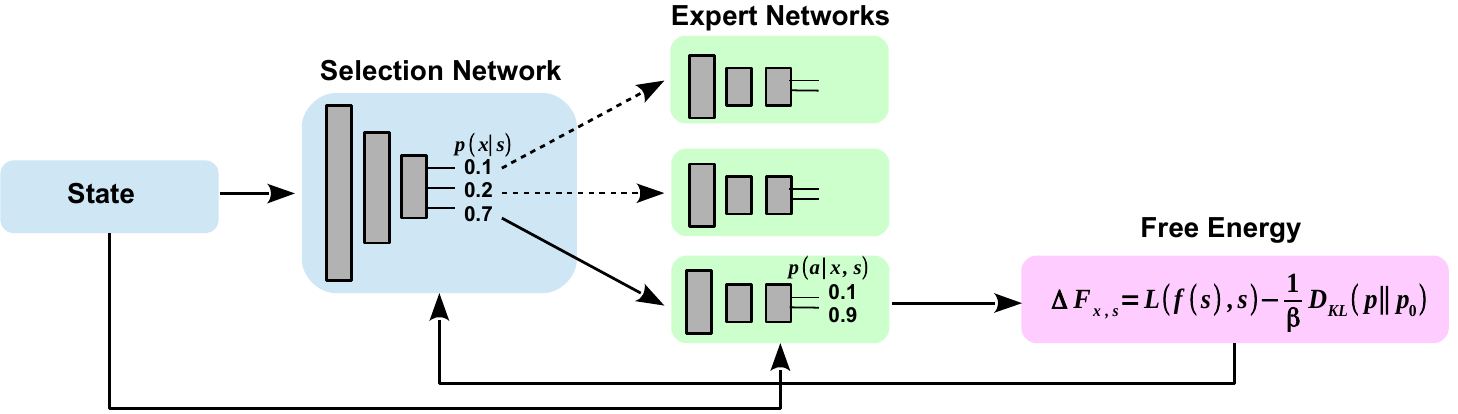}
\caption{The objective is optimized by dividing the problem set into $k$ soft partitions and training each expert $x_i$ on the assigned region. The solid lines represent the hierarchical decisions made while the dashed lines represent the possibilities. The experts are updated using the specific free energy $L(f(s),s) - \frac{1}{\beta}\DKL(p||p_0)$, where $s$ is the current state, $f(s)$ the expert's response in state $s$, and $p_0, p$ the expert's prior and posterior distribution.}
\label{fig:graphmodel}
\end{figure}

Combining several bounded-rational agents by a selection policy allows for solving optimization problems that exceed the capabilities of the individual decision-makers \cite{Genewein2015}. To achieve this, the search space is split into optimal partitions, that can be solved by the individual decision-makers. A two stage mechanism is introduced: The first stage is comprised of an expert selection policy $p(x|s)$ that chooses an expert $x$ based on the past performance of $x$ given the state $s$. This requires a high processing power in the expert selection, such that an optimal mapping from states to experts can be learned. The second stage chooses an action according to the expert's policy $p(a|s,x)$. The optimization problem given by \eqref{eq:br_obj} can be extended to incorporate a trade-off between computational costs and utility in both stages in the following way:
\begin{equation}
\label{eq:par_mutual}
\max_{p(a|s,x), p(x|s)} \E[\mathbf{U}(s,a)] - \frac{1}{\beta_1}I(S;X) - \frac{1}{\beta_2}I(S;A|X),
\end{equation}
where $\beta_1$ is the resource parameter for the expert selection stage and $\beta_2$ for the experts. $I(\cdot;\cdot)$ is the mutual information between the two random variables. The solution can be found by iterating the following set of equations \cite{Genewein2015}: 
\begin{equation}
\begin{cases}
\begin{array}{rcl}
p(x|s) &=& \frac{1}{Z(s)}p(x) \exp(\beta_1 \F(s,x)) \\[2pt]
p(x) &=& \sum_s p(s) p(x|s) \\[2pt]
p(a|s,x) &=& \frac{1}{Z(s,x)} p(a|x) \exp(\beta_2 \mathbf{U}(s,a)) \\[2pt] 
p(a|x) &=& \sum_s p(s|x)p(a|s,x) , \\[2pt]
\end{array}
\end{cases}
\end{equation}
where $Z(s)$ and $Z(s,x)$ are normalization factors and $\F(s,x) = \E_{p(a|s,x)}[\mathbf{U}(s,a)] - \frac{1}{\beta_2}\DKL(p(a|s,x)\vert\vert p(a|x))$ is the \textit{free energy} of the action selection stage. The effective distribution $p(a|s)$ encapsulates a \emph{mixture-of-experts} policy consisting  of the experts $p(a|s,x)$ weighted by the responsibilities $p(x|s)$. Note that the Bayesian posterior is not determined by a given likelihood model, but is the result of the optimization process~\eqref{eq:par_mutual}. The hierarchical system is depicted in Figure \ref{fig:graphmodel}.
\subsection{Maximum Entropy Reinforcement Learning}
We model sequential decision problems by defining a Markov Decision Process as a tuple $(\mathcal{S}, \mathcal{A}, P, r)$, where $\mathcal{S}$ is the set of states, $\mathcal{A}$ the set of actions, $P: \mathcal{S} \times \mathcal{A} \times \mathcal{S} \rightarrow [0,1]$ is the transition probability, and $r: \mathcal{S} \times \mathcal{A} \rightarrow \mathbb{R}$ is a reward function. The aim is to find the parameter $\theta$ of a policy $\pi_\theta$ that maximizes the expected reward:

\begin{equation}
\theta^* = \argmax_{\theta} \underbrace{\mathbb{E}_{\tau \thicksim \pi_\theta}\left[\sum_{t=0}^\infty r(s_t, a_t)\right]}_{J(\pi_\theta)}.
\end{equation}
We define $r(\tau) = \sum_{t=0}^\infty r(s_t, a_t)$ as the cumulative reward of trajectory $\tau = \{(s_t, a_t)\}_{t=0}^\infty$,  which is sampled by acting according to the policy $\pi$, i.e. $(s, a)\sim \pi(\cdot|s),$ and $s_{t+1} \sim P(\cdot|s_t, a_t)$.
Learning in this environment can then be modeled by reinforcement learning \cite{Sutton2018}, where an agent interacts with an environment defined by the tuple $(\mathcal{S}, \mathcal{A}, P, r)$ over a number of (discrete) time steps $t$. At each time step $t$, the agent receives a state $s_t$ and selects an action $a_t$ according to the policy $\pi(a_t|s_t)$. In return, the agent receives the next state $s_{t+1}$ and a scalar reward $r_t$. This process continues until the agent reaches a terminal state after which the process restarts. The goal of the agent is to maximize the expected return from each state $s_t$, which is typically defined as the infinite horizon discounted sum of the rewards. A common choice to achieving this is Q-Learning \cite{Watkins1992}, where we make use of an action-value function that is defined as the discounted sum of rewards:
\begin{equation}
Q(\tau) =\sum_{t=0}^\infty \gamma^t r(s_{t}, a_{t}),
\end{equation}
and $\gamma \in (0, 1]$ is a discount factor.
Learning the optimal policy can be achieved in many ways. Here, we consider Policy gradient methods \cite{Sutton2000} which are a popular choice to tackle reinforcement learning problems. The main idea is to directly manipulate the parameters $\theta$ of the policy in order to maximize the objective $J(\pi_\theta)$ by taking steps in the direction of the gradient $\nabla_\theta J(\pi_\theta)$. The gradient of the policy can be written as 
\begin{equation}
\nabla_\theta J(\pi_\theta) = \mathbb{E}_{\tau \sim \pi_\theta}\left[\nabla_\theta \log \pi_\theta(\tau)Q(\tau)\right],
\end{equation}
where $\pi_\theta(\tau) = p(s_0) \Pi_{t=0}^T\pi_\theta(a_t|s_t)p(s_{t+1}|s_t,a_t)$ is the probability of a trajectory $\tau$. This result of the policy gradient theorem has inspired several algorithms, which often differ in how they estimate the cumulative reward, e.g. Q-Learning \cite{Watkins1992},  the REINFORCE algorithm \cite{Williams1992}, and Actor-Critic algorithms \cite{Konda2000}. In this study we will introduce an hierarchical Actor-Critic algorithm.

To balance exploration vs. exploitation \cite{Sutton2000} maximum entropy reinforcement learning introduces an additional policy entropy term as a penalty to the value function. The optimal value function under this constraint is defined as 
\begin{equation}
\pi^*(a|s) = \max_{\pi}\mathbb{E}_{\pi}\left[\sum_{t=0}^\infty\gamma^t \left( r(s_t, a_t) - \frac{1}{\beta} \log \pi(a|s)\right)\right].
\end{equation}
Here, $\beta$ trades off between reward and entropy, such thath $\beta \rightarrow \infty$ recovers the standard RL value function and $\beta \rightarrow 0$ recovers the value function under a random policy.
One can define this objective as an inference problem \cite{levine2018reinforcement} by specifying a fixed prior distribution $\pi_0(a)$ over trajectories. In the next sections we generalize this assumption by assuming the prior distribution to be part of the optimization problem, as discussed in the earlier section on bounded rationality. 

\begin{algorithm}[t!]
\scriptsize
\caption{Specialization in Hierarchical Systems for Reinforcement Learning}
\label{alg:shs}
\begin{algorithmic}[1]
\State \textbf{Input}: Initial selection policy parameters $\theta$, initial free energy value function parameters $\phi$, initial expert policy parameters $\vartheta$, initial expert value function parameters $\varphi$, environment $E$, number of training episodes $K$, number trajectories per update step $N$\
\State \textbf{Hyperparameters}: penalty parameters $\beta_1$, $\beta_2$, learning rates $\alpha_s$, $\alpha_x$ for selector and experts, discount factor $\gamma$, prior momentum $\lambda$
\State Initialize prior parameters $\theta, \phi$
\For{$k$ = 0, 1, 2, ..., $K$}
\State Collect set of $N$ trajectories $\mathcal{D}_k = \{\tau_i\}$ by running policies $\pi_{\theta}(x|s)$ and $\pi_{\vartheta}(a|s,x)$ in environment $E$
\State Compute rewards-to-go $R_t$ and free energies-to-go $F_t$ with discount factor $\gamma$:
\begin{eqnarray*}
R_t &=& \sum_{l=0}^T\gamma^lr(s_{t+l}, a_{t+l}) \\
F_t &=& \sum_{l=0}^T\gamma^l f(s_{t+l}, x_{t+l}, a_{t+l}),
\end{eqnarray*}
where $f(s,x,a) = r(s, a) - \frac{1}{\beta_2} \log\frac{\pi_{\vartheta}(a|s,x)}{\pi(a|x)}$.
\State Estimate advantages $\hat{A}^F_t$ and $\hat{A}_t$ based on the current value functions $V_{\phi}$ and $V_{\varphi}$ as:
\begin{eqnarray*}
\hat{A}^F_t &=& r(s_t, a_t) + \gamma V_{\phi}(s_{t+1}) - V_{\phi}(s_t) \\
\hat{A}_t &=& f(s_t, x_t, a_t) + \gamma V_{\varphi}(s_{t+1}) - V_{\varphi}(s_t)
\end{eqnarray*}
\State Estimate expert policy gradients $\hat{g}_{x}$ (for each expert $x$) and selector policy gradients $\hat{g}_s$ as 
\begin{eqnarray*}
\hat{g}_x &=& \frac{1}{|\mathcal{D}_k|}\sum_{\tau \in \mathcal{D}_k}\sum_{t=0}^T\nabla_{\vartheta} \log \pi_{\vartheta}(a_t|x_t,s_t)\hat{A}_t \\
\hat{g}_s &=& \frac{1}{|\mathcal{D}_k|}\sum_{\tau \in \mathcal{D}_k}\sum_{t=0}^T\nabla_\theta \log \pi_\theta(x_t|s_t)\hat{A}^F_t
\end{eqnarray*}
\State Update policy parameters $\vartheta, \theta$ with gradients $\hat{g}_x, \hat{g}_s$
\State Fit value functions by regression:
$$\phi = \argmin_\phi \frac{1}{|\mathcal{D}|T}\sum_{\tau\in\mathcal{D}}\sum_{t=0}^T (V_{\phi}(s_t) - R_t)^2$$
$$\varphi = \argmin_\varphi \frac{1}{|\mathcal{D}|T}\sum_{\tau\in\mathcal{D}}\sum_{t=0}^T (V_{\varphi}(s_t) - F_t)^2$$
\State Update priors $\pi(a|x)$ and $\pi(x)$
\EndFor\
\State \textbf{return} $\theta$, $\vartheta$, $\phi$, $\varphi$
\end{algorithmic}
\end{algorithm}

\section{Specialization in Hierarchical Systems}
\label{sec:spec}
In this section we introduce our novel gradient based algorithm to learn the components of a hierarchical multi-agent policy. Information-theoretic bounded rationality argues that hierarchies and abstractions emerge when agents have only limited computational resources \cite{Genewein2015}. In particular, we leverage the hierarchical model introduced earlier to learn a disentangled representation of the state and action spaces. We will see how limiting the amount of uncertainty each agent can reduce leads to specialization. First we will show how this principle can be transformed into a general on-line learning algorithm and afterwards we will show how it can be applied to classification as an illustrative example and reinforcement learning. In the following we will derive our algorithm with a focus on reinforcement learning.

The model consists of two stages: an expert selection stage, and an action selection stage. The first stage learns a soft partitioning $\pi(x|s)$ of the state space and assigns each partition optimally to the experts according to a parametrized policy $\pi_\theta(x|s)$ with parameters $\theta$ such that $\sum_x \pi_\theta(x|s) \F(s,x)$ is maximized under an information-theoretic constraint on $\mathbb{E}\left[\DKL(\pi_\theta(x|s)||\pi(x))\right]$. The second stage is defined by a set of policies $\pi_\vartheta(a|s,x)$ that maximize the expected utility $\sum p(s|x)\sum p_\theta(a|s,x)\mathbf{U}(s,a)$ for each expert $x$. We device a gradient based on-line learning algorithm to find the optimal parameters in the following. 

Firstly, we note that in the reinforcement learning setup the utility $\mathbf{U}(s,a)$ is given by the reward function $r(s,a)$. And secondly that in maximum entropy RL the regularization penalizes deviation from a fixed equally distributed prior, but in a more general setting we can discourage deviation from an arbitrary prior policy by optimizing for: 

\begin{equation}
\argmax_{\pi}\mathbb{E}_{\pi}\left[\sum_{t=0}^\infty\gamma^t \left( r(s_t, a_t) -  \frac{1}{\beta} \log \frac{\pi(a_t|s_t)}{\pi(a)}\right)\right].
\end{equation}
As discussed in Section \ref{seq:br} the optimal prior (in terms of an optimal utility vs. processing cost) is the marginal of the posterior policy given by $\pi(a) = \sum_s p(s) \pi(a|s)$. We can incorporate this into the optimization problem \eqref{eq:par_mutual} by rewriting it to:
\begin{equation}
\label{eq:overall_objective}
\max_{\pi_\vartheta(a|s,x), \pi_\theta(x|s)} \sum_{s,x,a}p(s)  \pi_\theta(x\vert s) \pi_\vartheta(a\vert s,x)J(s,x,a)
\end{equation}
where the objective $J(s,x,a)$ is given by
\begin{figure}[t!]
\centering
\includegraphics[width=.5\textwidth, trim={4.0cm 1.35cm 3.5cm 1.60cm}, clip]{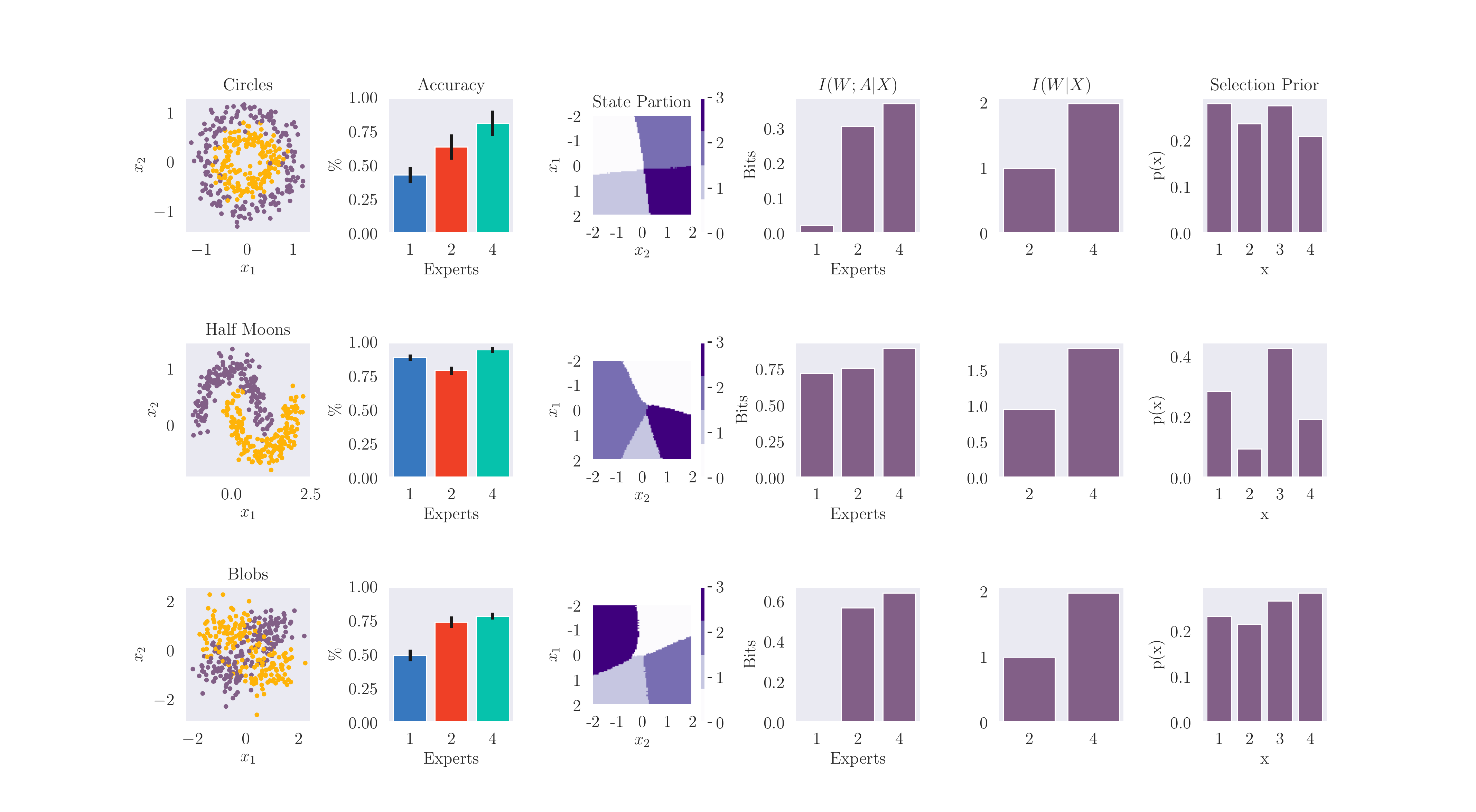}
\caption{Results for three synthetic classification tasks. Our system is successfully learning a partition of the sample space such that the linear experts are able to classify their assigned samples correctly. As expected the accuracy improves while mutual information also increases as we add experts. The last column shows the expert usage by the selection prior $p(x)$.}
\label{fig:classification}
\end{figure}
\begin{equation}
J(s,x,a) = \mathbf{U}(s,a) - \frac{1}{\beta_1}\log \frac{\pi_\theta(x\vert s)}{\pi(x)} - \frac{1}{\beta_2}\log \frac{\pi_{\vartheta}(a\vert s,x)}{\pi(a \vert x)},
\end{equation}
and $\theta, \vartheta$ are the parameters of the selection policy and the expert policies, respectively. Note that each expert policy has a distinct set of parameters $\vartheta_{x}$, i.e. $\vartheta = \{\vartheta_x\}_x$, but we drop the $x$ index for readability. Considering our algorithm is on-line and the action space is continuous, it would be prohibitive to compute the prior in each step. Instead we approximate the prior distributions $\pi(x)$ and $\pi(a|x)$ by exponential running mean averages of the posterior policies with momentum terms $\lambda_1$ and $\lambda_2$:
\begin{eqnarray}
\pi_{t+1}(a|x) &=& \lambda_1\pi_t(a|x) + (1 - \lambda_1)\pi_\vartheta(a|s,x) \\
\pi_{t+1}(x) &=& \lambda_2\pi_t(x) + (1 - \lambda_2)\pi_{\theta}(x|s)
\end{eqnarray}
In our experiments we set $\lambda_1, \lambda_2$ to 0.99.
To optimize the objective we define two separate value functions: one to estimate the discounted sum of rewards and one to estimate the free energy of the expert policies. The discounted reward for the experts is 
\begin{equation}
R_t = \sum_{l=0}^T\gamma^lr(s_{t+l}, a_{t+l}),
\end{equation}
which again we learn by parameterizing the value function with a neural network and performing regression on
\begin{equation}
\phi^* = \argmin_\phi \frac{1}{|\mathcal{D}|T}\sum_{\tau\in\mathcal{D}}\sum_{t=0}^T (V_{\phi}(s_t) - R_t)^2,
\end{equation}
where $\phi$ are the parameters of the value net and $\mathcal{D}$ is a set of trajectories $\tau$ up to horizon $T$ collected by roll-outs of the policies. Similar to the discounted reward $R_t$ we can now define the discounted free energy $F_t$ as
\begin{equation}
F_t = \sum_{l=0}^T\gamma^l f(s_{t+l}, x_{t+l}, a_{t+l}),
\end{equation}
where $f(s,x,a) = r(s, a) - \frac{1}{\beta_2} \log\frac{\pi_{\vartheta}(a|s,x)}{\pi(a|x)}$. The value function $F_t$ is learned by parameterizing the value function with a neural network and performing regression on the mean-squared-error:
\begin{equation}
\varphi^* = \argmin_\varphi \frac{1}{|\mathcal{D}|T}\sum_{\tau\in\mathcal{D}}\sum_{t=0}^T (V_{\varphi}(s_t) - F_t)^2,
\end{equation}
where $\varphi$ are the parameters of the value net, and $\mathcal{D}$ is a set of trajectories.

\subsection{Expert Selection} 
The selector network learns a policy $\pi_{\theta}(x\vert s)$ that maps states $s$ to expert policies $x$, based on their performance in the past. The resource parameter $\beta_1$ constrains how well this gating step can differentiate between the experts. For $\beta_1 \rightarrow 0$ the selection maps each state $s$ equally distributed to each experts or all to one single expert, depending on $\beta_2$. For $\beta_1 \rightarrow \infty$, $\pi_{\theta}(x\vert s)$ converges to the perfectly rational selector which always chooses the optimal expert $x$. Thus, the expert selection stage optimizes the following objective:
\begin{equation}
\max_\theta \E_{\pi_\theta(x|s)}\left[\hat{f}(s,x) - \frac{1}{\beta_1} \log\frac{\pi_{\theta}(x|s)}{\pi(x)}\right],
\end{equation}
where $\hat{f}(s,x) = \mathbb{E}_{\pi_\vartheta(a|s,x)}[f(s, x, a)]$, which is the free energy of the expert. The gradient of $J(\theta)$ is then given (up to an additive constant) by 
\begin{equation}
\E\left[\nabla_\theta \log \pi_\theta(x \vert s)\left(\hat{f}(s, x) - \frac{1}{\beta_1} \log\frac{\pi_{\theta}(x|s)}{\pi(x)}\right) \right].
\label{eq:selgrad}
\end{equation}
The double expectation can be replaced by Monte Carlo estimates, where in practice we use a single $(s,x,a)$ tuple for $\hat{f}(s,x)$.
This formulation is known as the policy gradient method \cite{Sutton2000} and is prone to producing high variance gradients. A common technique to reduce the variance is to formulate the updates using the advantage function instead of the reward \cite{schulman2015high,mnih2016asynchronous}. The advantage function $A(a_t,s_t)$ is a measure of how well a certain action $a$ performs in a state $s$ compared to the average performance in that state, i.e. $A(a,s) = Q(s,a) - V(s)$. Here, $V(s)$ is called the value function and is a measure of how well the agent performs in state $s$, and $Q(s,a)$ is an estimate of the cumulative reward achieved in state $s$ when the agent executes action $a$. Thus the advantage is an estimate of how advantageous it is to pick $a$ in state $s$ in relation to a baseline performance $V(s)$. Instead of learning the value and the Q function, we can approximate the advantage function in the following way:
\begin{equation}
\hat{A}^F_t = \underbrace{f(s_t, x_t, a_t) + \gamma V_{\phi}(s_{t+1})}_{\approx Q(s_t, a_t)} - V_{\phi}(s_t).
\end{equation}
This yields the following gradient estimates for the selector:
\begin{equation}
\hat{g}_s = \frac{1}{|\mathcal{D}|}\sum_{\tau \in \mathcal{D}}\sum_{t=0}^T\nabla_\theta \log \pi_\theta(x_t|s_t)\hat{A}^F_t,
\end{equation}
where $\mathcal{D}$ is a set of trajectories $\tau$ produced by the policies. This formulation allows us to perform the updates as in Advantage-Actor-Critic-Models. The expectation can be estimated via Monte Carlo sampling which enables us to employ our algorithm in an on-line optimization fashion.

\subsection{Action Selection} 
The actions $a$ are given by the posterior action distribution of the experts. Each expert $x$ maintains a policy for each of the world states $s$ and updates those according to the utility/cost trade-off. The advantage function for each expert is given as
\begin{equation}
\hat{A}_t = r(s_t, a_t) + \gamma V_{\varphi}(s_{t+1}) - V_{\varphi}(s_t).
\end{equation}
The objective of this stage is then to maximize the expected advantage $\E_{p(a|s,x)}\big[\hat{A}_t\big]$, yielding the gradient estimates
\begin{equation}
\hat{g}_x = \frac{1}{|\mathcal{D}|}\sum_{\tau \in \mathcal{D}}\sum_{t=0}^T\nabla_\vartheta \log \pi_\vartheta(a_t|x_t,s_t)\hat{A}_t
\end{equation}
for each of the experts.

The algorithm we propose to find such an optimal hierarchical structure is based on the alternating optimization paradigm \cite{Bezdek2003}. During one phase the expert selector distribution $\pi(x|s)$ is optimized while the action selectors are held fixed and vice versa. The length of the phases can either be fixed or limited by checking for a convergence criterion. The complete algorithm is given in Algorithm \ref{alg:shs}.

\section{Experiments and Results}
In the following we will show how our approach can be applied to learning tasks where the overall complexity of the problem exceeds the processing power of (linear) experts. In particular, we will look at classification, regression, non-linear control (gain scheduling), and reinforcement learning. In our experiments we model the selector and both value functions as artificial neural networks and train them according to the gradient estimates we derived in Section \ref{sec:spec}.

\subsection{Classification and Regression with Linear Decision-Makers}

\begin{figure}[t!]
\centering
\includegraphics[width=0.35\textwidth, trim={.0cm .1cm 1cm 0.4cm}, clip]{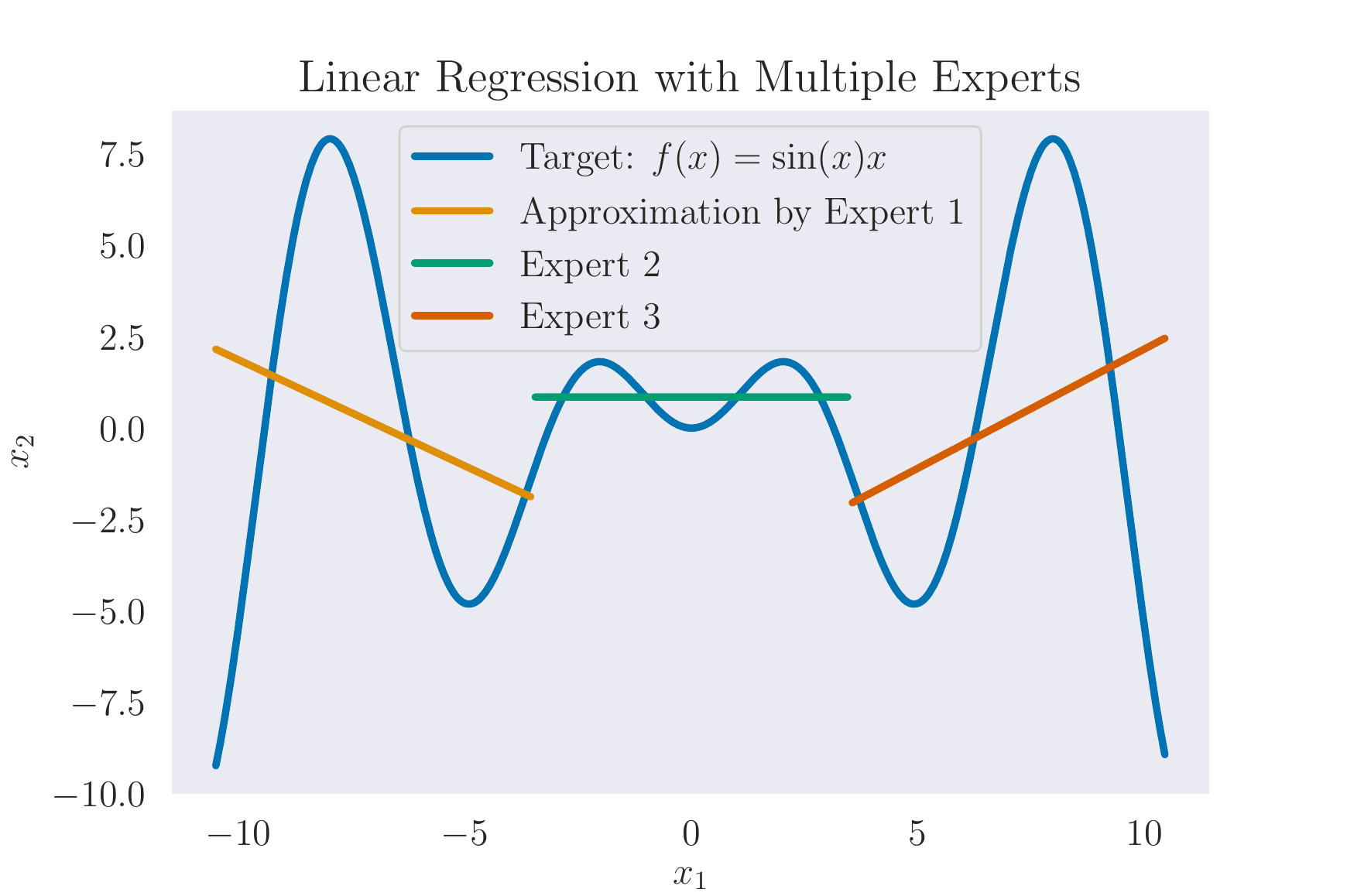}
\caption{Results for regression. Our system is successfully learning a partition of the sample space such that the linear experts are able to regress the non-linear function. }
\label{fig:regression}
\end{figure}

When dealing with complex data for classification (or regression) it is often beneficial to combine multiple classifiers (or regressors). In the framework of ensemble learning, for example, multiple classifiers are joined together to stabilize learning and improve the results \cite{Kuncheva2004}, such as Mixture-of-Experts \cite{Yuksel2012} and Multiple Classifier systems \cite{Bellmann2018}. The method we propose when applied to classification problems can be interpreted as a member of this family of algorithms. The application to regression could be considered as an example of local linear regression \cite{atkeson1997locally}. In accordance with Section \ref{seq:br} we define the utility as the negative loss, i.e. $U(\hat{y}, y) = -\mathcal{L}(\hat{y},y)$, where $\hat{y}$ is the action of the expert, i.e. the estimated class label or the estimated regression value, and $y$ is the ground truth, and $s$ are the input features. For classification, we chose the cross-entropy loss $\mathcal{L}(y, \hat{y}) = \sum_i y_i \log \frac{1}{\hat{y}_i} = -\sum_i y_i \log \hat{y}_i$ as a  performance measure, for regression the mean squared error $\mathcal{L}(y, \hat{y}) = \sum_i (\hat{y}_i-y_i)^2 $. The objective for expert selection becomes 
\begin{equation}
\max_\theta \E_{p_\theta(x|s)}\left[\hat{f}(x,s) - \frac{1}{\beta_1}\log\frac{p_\theta(x|s)}{p(x)}\right],
\end{equation}
where $\hat{f}(x,s) \coloneqq \mathbb{E}_{p_\vartheta(\hat{y}|x,s)}\big[-\mathcal{L}(\hat{y},y) - \frac{1}{\beta_2}\log\frac{p_\vartheta(\hat{y}|s,x)}{p(\hat{y}|x)}\big]$, i.e. the free energy of the expert $x$. For action  selection the objective then becomes
\begin{equation}
\max_\vartheta \E_{p_\vartheta(\hat{y}|x,s)}\left[-\mathcal{L}(\hat{y},y) - \frac{1}{\beta_2}\log\frac{p_\vartheta(\hat{y}|s,x)}{p_(\hat{y}|x)}\right].
\end{equation}
To find an optimal partitioning we consider the limit $\beta_1\rightarrow\infty$. We evaluated our method on three synthetic datasets for classification---see Figure \ref{fig:classification}---, and one synthetic dataset for regression---see Figure~\ref{fig:regression}. Our method is able to partition the problem set into subspaces and fit a linear decision-maker on each subset. This is achieved by the emergence of a hierarchy of specialized agents, as is evidenced by the emerged state partitioning. We implemented our experiments using TensorFlow \cite{Abadi2016} and scikit-learn \cite{scikit-learn}.

\subsection{Gain Scheduling by Combining Linear Decision-Makers}

When dealing with non-linear dynamics one approach is to decompose it into several linear sub-problems and design linear controllers for each sub-problem. This method is known as gain scheduling and is a well established method in the control literature \cite{rugh2000research}. In some cases it is possible to find auxiliary variables that correlate well with the changes in the underlying dynamics. It is then possible to reduce the effects of the parameter variations simply by changing the parameters of the regulator as a function of the auxiliary variables. For example, in flight control systems the Mach number and the dynamic pressure are measured by sensors and used as scheduling variables. A main problem in designing such systems is to find suitable scheduling variables. This is usually done by incorporating prior knowledge about the system dynamics. Here, we demonstrate how our approach can be applied to automatically learn the scheduling regions and pertinent linear controllers without pre-specifying the operating points. As an illustrative example, consider a scalar non-linear plant defined by the following piecewise linear dynamics:
\begin{equation}
\dot{x} = A_ix + B_iu + \epsilon, \qquad \text{for } x \in X_i
\end{equation}
where $x$ is the system state, $A_i$ are the system matrices, $B_i$ are the control matrices and $\epsilon$ is a random Gaussian noise source. Here, $\{X_i\}$ is a state partition into piecewise linear (affine) control regimes.  To approach this problem, we denote the plant state as states $s$, the control signal $u$ as the action and learn a set of linear Gaussian control policies (i.e. the experts), where we perform gradient descent to find the optimal parameters, as described in Section \ref{sec:spec}. 

\begin{figure}[t!]
\centering
\includegraphics[width=.5\textwidth, trim={3cm 0cm 3cm 0.9cm}, clip]{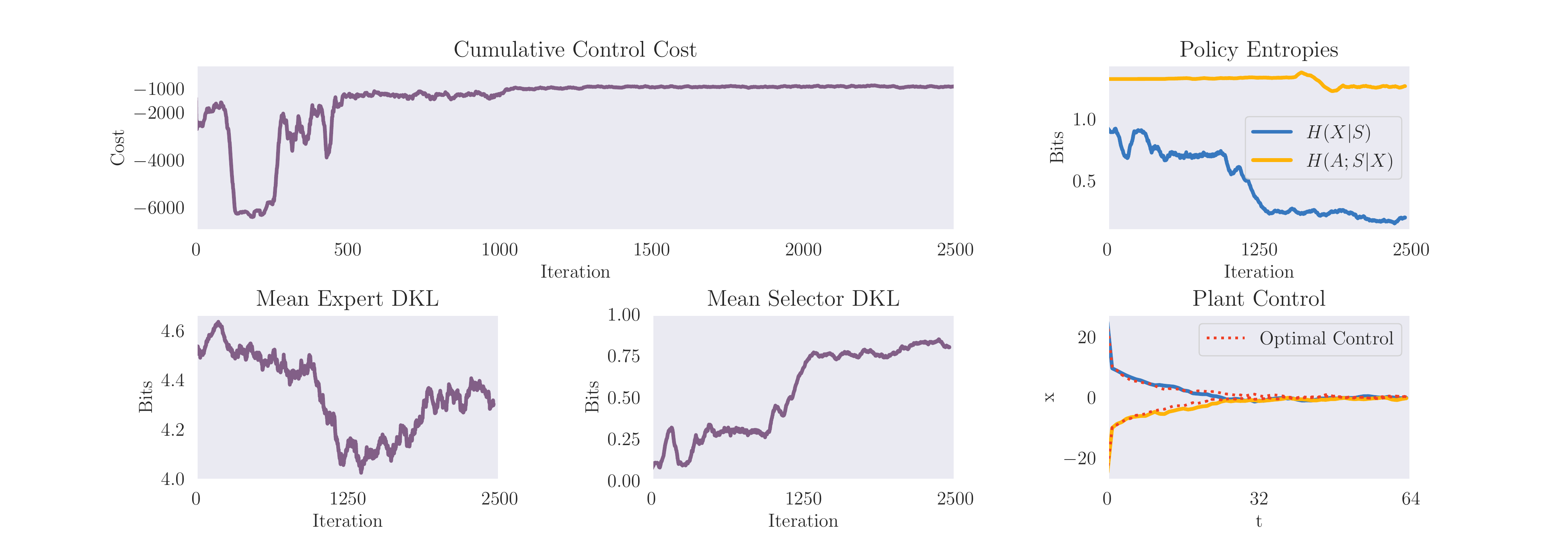}
\caption{The plant is controlled by finding the optimal partitions $X_i$, which allow the hierarchical control system to steer the plant towards the stationary point at zero, as depicted in upper right figure (solid lines). The dashed lines represent the system state when controlled by hand-tailored gain scheduling. Importantly, in our algorithm the scheduling policy is not given but the result of an optimization process. We ran the optimization for 2500 iterations with 64 time steps each.}
\label{fig:gain}
\end{figure}

Consider the following operation regimes:
\begin{equation}
B_i =
\begin{cases}
1 & \text{if } x \in X_0  \\
-1  & \text{if } x \in X_1,
\end{cases}
\end{equation}
where $X_0 = \{x|x \geq 0\}$ and $X_i = \{x|x < 0\}$. Assuming quadratic costs $J = \frac{1}{2}\sum_{t=0}^T Qx_t^2 + Ru_t^2$, it is straightforward to find an optimal controller for each partition $X_i$ and we can switch between the regimes by defining the system state as the scheduling variable. We set $Q = 1.0$ and $R = 0.01$ and $T=64$ and find the optimal gains $K_0 = 11.05$ and $K_1 = -11.05$, for regimes $X_0$ and $X_1$ respectively.
Obviously, this is only possible if the plant dynamics are known. In contrast, our algorithm is able to learn the scheduling policy and the control policy automatically, which is shown in Figure \ref{fig:gain}. The gains found by our algorithm were $K_0 = 9.99$ and $K_1 = -9.71$ and achieve a control cost of $J = -934$ compared to $J = -1004$ achieved by gain scheduling. We set the noise source to $\epsilon \thicksim \mathcal{N}(0, 0.25)$, causing the control policy to shift when the plant state is close to zero. The prior is $p(x) = (0.55, 0.45)$, showing that both expert policies were used to control the plant. The mutual information of $I(S;X) \approx 0.85$ bits and the entropy of $p(x|s)$ decaying towards 0 show that the selector successfully learns to partition the state space. The resource parameter was set to $\beta_2 = 0.001$ to drive specialization.

\subsection{Reinforcement Learning with Linear Decision-Makers}

\begin{figure*}[t!]
\centering
\includegraphics[width=0.95\textwidth, trim={3cm 0.1cm 3cm .5cm}, clip]{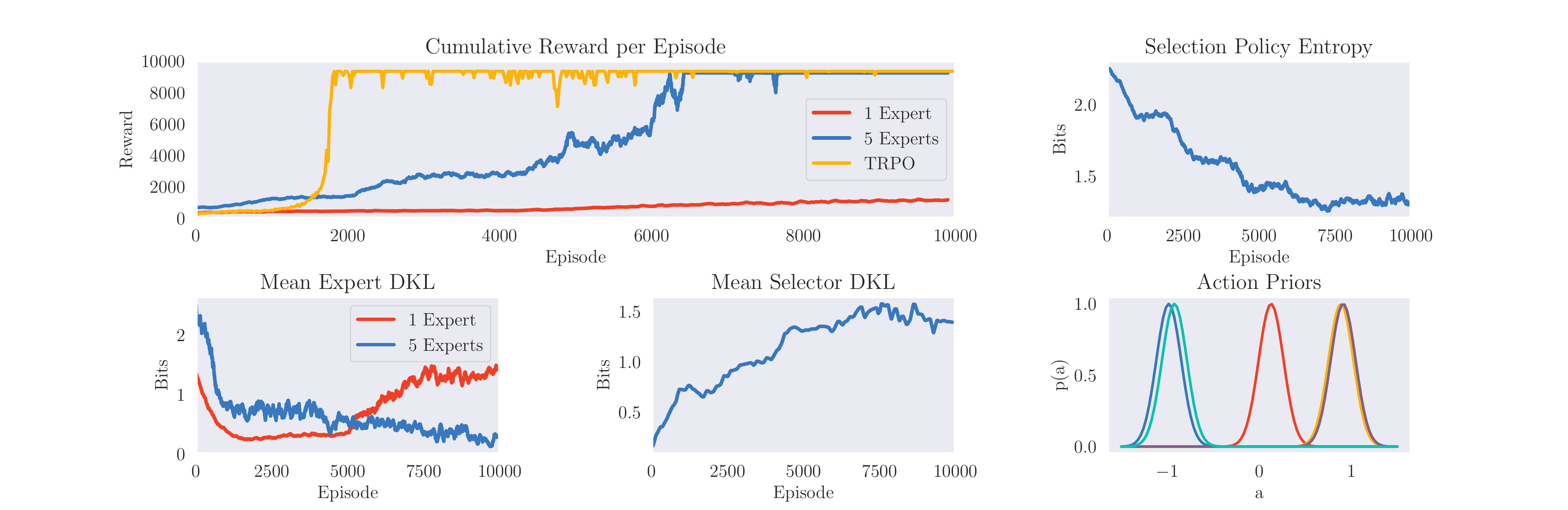}
\caption{Results for the inverted double pendulum problem. The upper row shows the reward per episodes achieved by different systems. We show the performance of a system with one linear expert, five linear experts, and compare it to Trust Region Policy Optimization (TRPO) \cite{Schulman2015} (discussed further in Section \ref{sec:disc}).}
\label{fig:reinforcement}
\end{figure*}

In the following we will show how our hierarchical system is able to solve a continuous reinforcement learning problem using an optimal arrangement of linear control policies. We evaluate on a task known as Acrobot \cite{Sutton1996}, more commonly referred to as the inverted double pendulum. The task is to swing  a double-linked pendulum up and keep it balanced as long as possible. The agent receives a reward of 10 plus a distance penalty between its actual state and the goal state. The episode is terminated if the agent reaches a predefined terminal state (hanging downwards) or after 1000 time steps. To balance the pendulum the agent is able to apply a force to the central joint of $a \in [-1, +1]$, i.e. move it to the left or the right, respectively. This environment poses a non-linear control problem and can thus not be solved optimally by a single linear controller. We show that using our approach, a committee of linear experts can solve this non-linear task. The results are shown in Figure \ref{fig:reinforcement}. We allowed for five experts (both with $\beta_2 = 0.001$), but our system learns that three experts are sufficient to solve the task. The priors for each expert (lower right Figure, each color represents an expert) center on -1, 0, and 1, which correspond to swinging the double pendulum to the left, no force, and swinging to the right, respectively. The remaining two experts overlap accordingly. We can see that the average $\DKL(\pi_\vartheta(a|x,s)||\pi(a|x))$ in the five expert setup decreases, while the selection $\DKL(\pi_\theta(x|s)||\pi(x))$ increases to $\approx 1.5$. Both indicate that the system has learned an optimal arrangement of three experts and is thus able to achieve maximum reward and eventually catches up to the performance of TRPO, which is our nonlinear control baseline that was trained with three-layered policy and value neural networks consisting of ReLu activation functions. Our method successfully learned a partitioning of the double-pendulum state space without having any prior information about any of the system dynamics or the state space. We implemented our experiments in TensorFlow \cite{Abadi2016} and OpenAI Gym \cite{Brockman2016}.

\section{Discussion}
\label{sec:disc}
Recently, there has been increased interest in investigating the effects of information-theoretic constraints on reinforcement learning tasks with mixture-of-experts policies. For example, the authors of \cite{Ghosh2017} have proposed a divide-and-conquer principle for policy learning in a reinforcement learning setting. They argue that splitting a central policy into several sub-policies improves the learning phase by requiring less samples overall. To implement this idea they split the action and state space into \emph{pre-defined} partitions and train policies on these partitions. The information-theoretic constraints during training enforce that multiple experts are kept similar to each other,  so that the expert policies can be fused into one central policy. In contrast, in our approach the information-theoretic constraints enforce that all experts stay close to their respective priors thereby generating as little informational surprise as possible. This leads to specialization of experts, because each expert is assigned a sub-space of the input space where information can be processed efficiently without deviating too much from the optimal prior adapted to that region. Crucially, in our setup the partitioning is not predefined but part of the optimization process.

Our approach belongs to a wider class of models that use information constraints for regularization to deal more efficiently with learning and decision-making problems~\cite{Daniel2012,Neumann2013,Martius2013,Leibfried2015,Grau-Moya2016,Peng2017,Leibfried2017,Grau-Moya2017,Ghosh2017,Achiam2017,Hihn2018,Schach2018,grau2018soft,
Gottwald2019,gottwald2019bounded}. One such prominent approach is Trust Region Policy Optimization (TRPO) \cite{Schulman2015}. The main idea is to constrain each update step to a trust region around the current state of the system. This region is defined by the Kullback-Leibler Divergence $\DKL(\pi_\text{new}||\pi_\text{old})$ between the old policy and the new policy. The smooth updates provide a theoretic monotonic policy improvement guarantee. Another similar approach are relative entropy policy search methods \cite{Daniel2012}, where the idea is to learn a gating policy that can decide which sub-policy to choose. To achieve this the authors impose a $\DKL$ constraint between the data distribution and the next policy. 

Another related approach to ours are Mixture of Experts (ME) models, originally introduced by \cite{Jacobs1991} as a tree structure to solve complex classification and regression tasks by leveraging the divide and conquer paradigm. MEs consist of three main components: gates, experts, and a probabilistic model that combines the expert predictions. The objective of the gate is to find a soft partitioning of the input space and assign partitions to experts which perform best on the partition. Experts are built to perform optimally in regression or classification tasks given an assigned partition. The model is a weighted sum of the experts outputs, weighted by how confident the gate is in the experts opinion. MEs exhibit a high degree of flexibility as evidenced by the variety of models and algorithms employed in the three components \cite{Yuksel2012}. Our model allows learning such models, but can also be applied to more general decision-making scenarios like reinforcement learning.

Prior work to ours in the control literature working on a similar setup  assumed the system dynamics to be given. e.g. in \cite{Abramova2012,Randlov2000,Yoshimoto2005}. The authors of \cite{Abramova2012} split the state space of the inverted pendulum into predefined bins and find a linear controller that stabilizes each bin by learning a selection policy over these predefined controllers. Our approach differs from these by the fact that we take only the reward signal and perform a partitioning of the state space and learning linear controller on these partitions simultaneously. This poses a difficult learning problem as both system parts have to adjust to one another on different timescales. Other decentralized approaches (e.g. \cite{allamraju2017communication}) have trained separate decentralized models to fuse them into a single model that can be used by a reinforcement learning agent. In contrast, our method learns several sub-policies.

In summary, we introduce a promising novel gradient based on-line learning paradigm for hierarchical multi-agent systems. Our method finds an optimal soft partitioning by considering the agents' limitations in terms of information-theoretic constraints, supporting expert specialization. Importantly, our model is capable of doing so without any prior information about the task. This becomes especially difficult in continuous control tasks, where the system dynamics are unknown. Our method is abstract and principled in a way that allows it to be employed on a variety of tasks including multi-agent decision-making, mixture-of-expert regression, and divide-and-conquer reinforcement learning.
An open questions remains how to apply our method to high dimensional control tasks.

\bibliographystyle{plain}
\bibliography{bibliography.bib}


%
\end{document}